\documentclass{article}
\ifdefined\pdfsuppressptexinfo\pdfsuppressptexinfo=15\relax\fi
\usepackage{iclr2027_conference,times}

\usepackage{amsmath,amsfonts,bm}

\def\eqref#1{equation~\ref{#1}}

\def\1{\bm{1}}

\DeclareMathAlphabet{\mathsfit}{\encodingdefault}{\sfdefault}{m}{sl}
\SetMathAlphabet{\mathsfit}{bold}{\encodingdefault}{\sfdefault}{bx}{n}

\usepackage{hyperref}
\hypersetup{hidelinks}
\usepackage{url}
\usepackage{booktabs}
\usepackage{graphicx}
\usepackage{float}
\usepackage{flafter}
\usepackage{placeins}
\usepackage{multirow}
\usepackage{array}
\usepackage{amsmath}
\usepackage{amssymb}

\usepackage{xcolor}
\usepackage{tikz}
\usetikzlibrary{positioning,arrows.meta}

\title{REMEDY: How Far Is Video Generation from Medical Education World Models?}

\author{Lixing Tan$^{1}$, Yanghao Zhou$^{2}$, Qing Xia$^{1}$,Yuting Guo$^{3}$, Shuai Li$^{1}$, Aimin Hao$^{1}$\\
$^{1}$Beihang University\\
$^{2}$Beijing Institute of Technology\\
$^{3}$Beijing Information Science and Technology University}

\begin{document}

\maketitle

\begin{abstract}
Recent video generation models produce realistic videos and show potential
as a foundation for world models. These advances create opportunities for generating medical teaching demonstrations, which requires both convincing visual quality and precise procedural actions. However, whether current generators can meet these requirements has not been measured. To address this problem, we
introduce Readiness Evaluation of Medical Education Demonstration sYnthesis
(REMEDY), to our knowledge, the first benchmark for AI-generated medical
teaching demonstrations. REMEDY provides
900 first frames from real demonstration videos, covering 12 tasks across four
scenarios: operating room, imaging, clinic and bedside, and
resuscitation. Five contemporary open-source video generation models produce
4{,}500 videos from these frames. We combine task-specific clinical checklists
with video and motion quality metrics. Evaluation covers four dimensions: clinical
action following, clinical profiles, video quality, and motion quality.
Our results show that realistic appearance and temporal
consistency do not ensure correct clinical actions. Even the most advanced MiniMax-H3 achieves
only 28.25\% on strict clinical success rate, and fine-grained
clinical actions remain challenging.
These findings establish a foundation and roadmap for developing future medical
education world models.
\end{abstract}

\section{Introduction}
\label{sec:intro}
Recent Text-Image-to-Video (TI2V) models have improved in visual realism, temporal coherence, and motion quality. Given a first-frame
image and a text prompt, they can generate realistic videos with coherent
actions~\citep{wan2025,ltx2paper,li2026mtavg,zhou2026mtavg,zhou2026exoactor}. These capabilities support the
development of world models, extending video generation from synthesizing
images to simulating how environments change over time~\citep{cosmos2025,yuan2026helios}.
Medical research has begun to explore the potential of video generation.
For example, Endora uses video generation to simulate endoscopic scenes.
It generates high-quality endoscopic videos and explores their use in
downstream video analysis and the generation of 3D medical scenes with
consistent views~\citep{endora2024}. SurgVeo focuses on surgical
video generation within the endoscopic view. Surgeons assess the generated
videos in terms of visual appearance, instrument manipulation, environmental
feedback, and surgical intent. Its findings show that the evaluated models
can produce realistic surgical scenes but struggle with plausible
execution~\citep{surgveo2025}. MedGen covers a broader range of medical
scenarios. It fine-tunes a general-purpose video generator on medical videos
from multiple scenarios with detailed captions, improving visual quality and medical
accuracy~\citep{medgen2025}. However, its Text-to-Video (T2V) setting does
not use a real first frame to constrain people, instruments, or their spatial
relationships. It therefore cannot generate a continuation that follows
procedural requirements from a real clinical starting state.

We focus on generating medical teaching demonstrations from real clinical
first frames, with the performer in view. Compared with endoscopic surgical
videos that focus on a local operative field, these demonstrations center
on the performer and their interactions with patients and instruments.
These videos need to clearly show hand
placement, body posture, the sequence of steps, and responses to contact.
Generating such videos could provide demonstrations that can be viewed
repeatedly for clinical training, offer controllable visual settings for
surgical simulation, and help patients understand medical procedures.
These applications require accurate depictions of standard medical
procedures. Current models can generate natural, fluid human movements and
realistic scenes. But do they correctly perform the essential steps and
show the expected effects of the procedure? This question has not yet been
systematically evaluated.

Existing general-purpose video generation benchmarks are insufficient to
determine whether generated videos correctly demonstrate medical procedures.
General-purpose benchmarks such
as VBench and Video-Bench cover video quality, semantic alignment, and
action consistency, while UI2V-Bench further examines semantic understanding
and reasoning conditioned on an input image~\citep{vbench2024,videobench2025,ui2vbench2025}.
These assessments help determine whether the appearance and content meet
expectations, but do not by themselves establish procedural correctness.
Meanwhile, HumanScore evaluates human motion through kinematic plausibility,
temporal stability, and biomechanical consistency~\citep{humanscore2026}.
Its metrics, however, do not define procedure-specific criteria for essential
medical actions or their visible effects. As summarized in Table~\ref{tab:comparison},
existing benchmarks do not jointly assess clinical action following,
clinical profiles, video quality, and motion quality in medical
demonstrations with the performer in frame.

\begin{table}[!t]
\caption{\textbf{Comparison of REMEDY with existing video generation benchmarks.}
Benchmarks are compared by generation setting, clinical data, and evaluation
dimensions. $\checkmark$~=~full coverage, $\triangle$~=~partial coverage,
--~=~absent.}
\label{tab:comparison}
\centering
\scriptsize
\setlength{\tabcolsep}{2.5pt}
\begin{tabular*}{\linewidth}{@{\extracolsep{\fill}}llccccccc@{}}
\toprule
& & \multicolumn{3}{c}{\bf Clinical data} & \multicolumn{4}{c}{\bf Evaluation dimensions} \\
\cmidrule(lr){3-5}\cmidrule(lr){6-9}
\bf Benchmark & \bf Task &
\shortstack{\bf Medical\\\bf tasks} &
\shortstack{\bf Real clinical\\\bf frames} &
\shortstack{\bf Performer\\\bf in frame} &
\shortstack{\bf Clinical action\\\bf following} &
\shortstack{\bf Clinical\\\bf profiles} &
\shortstack{\bf Video\\\bf quality} &
\shortstack{\bf Motion\\\bf quality} \\
\midrule
VBench~\citep{vbench2024}          & T2V  & -- & -- & -- & -- & -- & $\checkmark$ & -- \\
Video-Bench~\citep{videobench2025} & T2V  & -- & -- & -- & -- & -- & $\checkmark$ & -- \\
UI2V-Bench~\citep{ui2vbench2025}   & I2V  & -- & $\triangle$ & -- & -- & -- & $\checkmark$ & -- \\
HumanScore~\citep{humanscore2026}  & T2V  & -- & -- & $\triangle$ & -- & $\triangle$ & -- & $\triangle$ \\
EduVQA~\citep{eduvqa2026}          & T2V  & -- & -- & -- & $\triangle$ & -- & $\triangle$ & -- \\
Endora~\citep{endora2024}          & T2V  & $\checkmark$ & -- & -- & -- & -- & $\checkmark$ & -- \\
Bora~\citep{bora2024}              & T2V  & $\checkmark$ & -- & -- & -- & -- & $\checkmark$ & -- \\
Med-VBench~\citep{medgen2025}      & T2V  & $\checkmark$ & -- & -- & -- & -- & $\checkmark$ & -- \\
SurgVeo~\citep{surgveo2025}        & I2V  & $\checkmark$ & $\checkmark$ & -- & -- & $\checkmark$ & $\checkmark$ & -- \\
\midrule
REMEDY (ours) & TI2V & $\checkmark$ & $\checkmark$ & $\checkmark$ & $\checkmark$ & $\checkmark$ & $\checkmark$ & $\checkmark$ \\
\bottomrule
\end{tabular*}
\end{table}

\begin{figure}[htb]
\centering
\includegraphics[width=\linewidth]{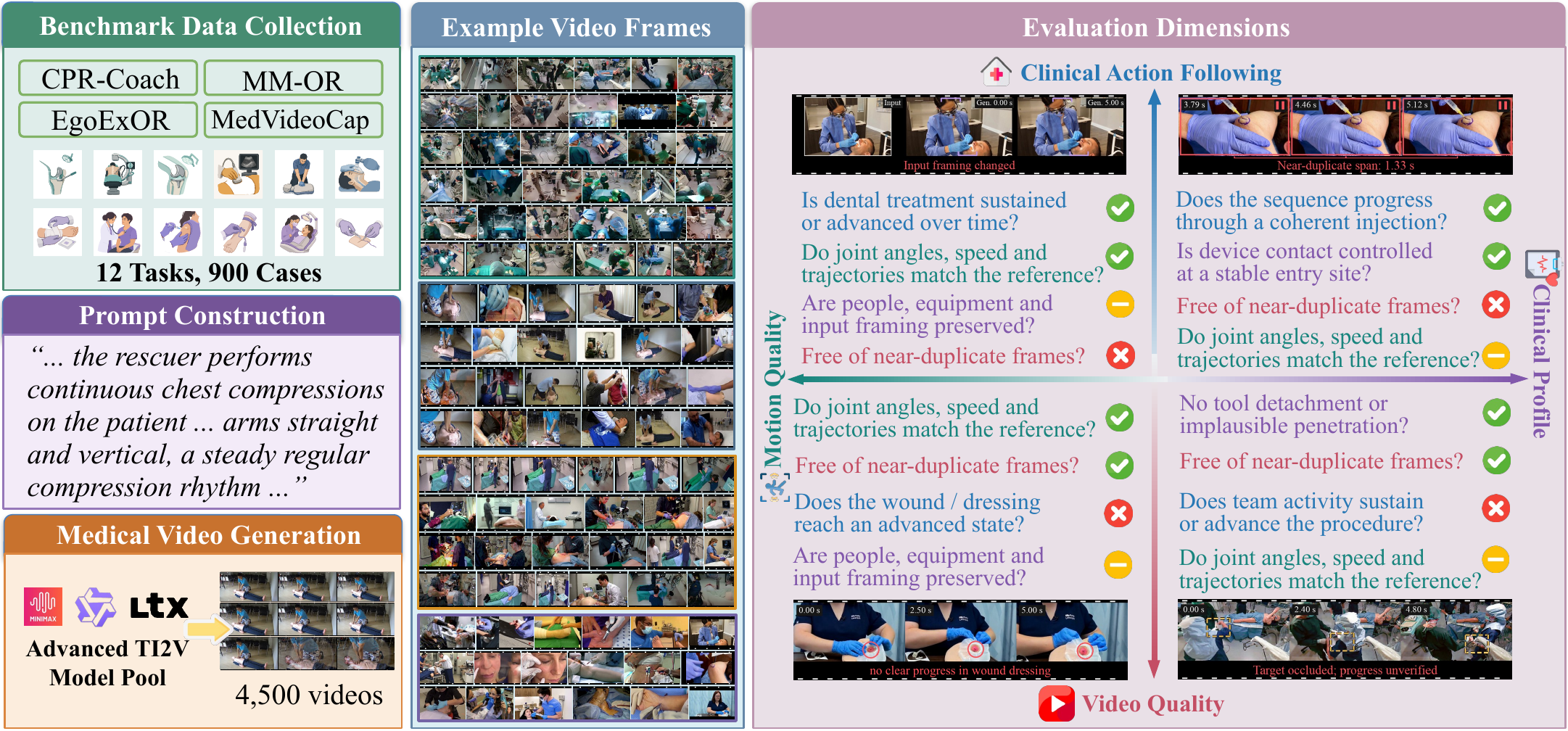}
\caption{\textbf{Overview of the REMEDY benchmark.}
Left: data collection, prompt construction, and video generation.
Middle: example videos across four medical scenarios.
Right: case-specific questions illustrate clinical action following,
clinical profiles, video quality, and motion quality. Green, yellow, and
red marks indicate satisfied, partly satisfied, and unsatisfied criteria.}
\label{fig:running}
\end{figure}

To address this gap, we introduce REMEDY, to our knowledge, the first
benchmark dedicated to AI-generated medical teaching demonstrations with
the performer in view, as shown in Figure~\ref{fig:running}. REMEDY contains
900 real clinical cases covering 12 tasks across four medical scenarios:
operating room, imaging, clinic and bedside, and resuscitation. Each case
includes a real first frame, a generation prompt, a real reference video,
and a clinical checklist. Given the same first frame and text prompt,
five contemporary open-source TI2V models produce 4{,}500 generated videos. We evaluate these videos
along four dimensions: clinical action following, clinical profiles, video
quality, and motion quality. On 450 human-reviewed generated videos, we also assess agreement
between human raters, between GPT-4o and humans, and between GPT-4o and other judges.

Our contributions are:
\begin{itemize}
  \item We formulate clinical failure diagnosis as a distinct evaluation
        problem for generated medical teaching demonstrations, assessing
        essential actions, clinical execution, and visible interaction effects
        beyond visual realism and plausible motion.
  \item We introduce REMEDY, with 900 real clinical cases spanning four medical
        scenarios and twelve tasks, and 4{,}500 generated videos. It provides
        shared inputs and clinical references for model comparison and failure
        analysis (Section~\ref{sec:bench}).
  \item We design a twelve-metric framework covering clinical action following,
        clinical profiles, video quality, and motion quality. Clinical checklists,
        frame evidence, video metrics, and reference-based motion comparisons
        distinguish clinical failures from visual and motion problems
        (Section~\ref{sec:rubric}).
  \item We evaluate five contemporary open-source TI2V models and find that even
        the strongest cannot reliably complete key actions across tasks.
        Action omissions and tool-tissue interaction failures persist despite
        visually convincing outputs (Sections~\ref{sec:main_results}
        and~\ref{sec:analysis_findings}).
\end{itemize}

\section{The REMEDY Benchmark}
\label{sec:bench}

To systematically evaluate the ability of contemporary open-source video
generation models to produce medical teaching demonstrations, we construct
the REMEDY benchmark.
Beyond visual realism and plausible motion, REMEDY focuses on
clinical correctness in medical teaching demonstrations: whether the
required actions are completed, procedural standards are followed, and
the expected visible effects occur. We use real reference videos to design
task-specific clinical checklists and link item-level judgments to frame
evidence. Together with independently computed video quality and motion
metrics, these judgments support model comparison and identify specific
unmet procedural requirements.

\subsection{Task definition}
\label{sec:task}

In medical teaching demonstration generation, a video model receives a real
clinical first frame and a procedure-specific prompt to generate a subsequent
action clip. The first frame provides the people, instruments, and spatial
arrangement that condition generation. The prompt specifies the clinical
actions, intended end state, and camera constraints. The model should depict
the required key actions and their expected visible effects while preserving
scene consistency.

To comprehensively evaluate video generation models, REMEDY assesses their
capabilities along four dimensions. Clinical action following measures
key-action completion through checklist decisions and frame evidence.
Clinical profiles assess clinical technique,
biomechanics, professionalism, and instruction following. Video quality
measures visual appearance and temporal consistency, while motion quality
compares body motion with real references.

\subsection{Data construction}
\label{sec:frames}

\begin{figure}[htb]
\begin{center}
\includegraphics[width=\linewidth]{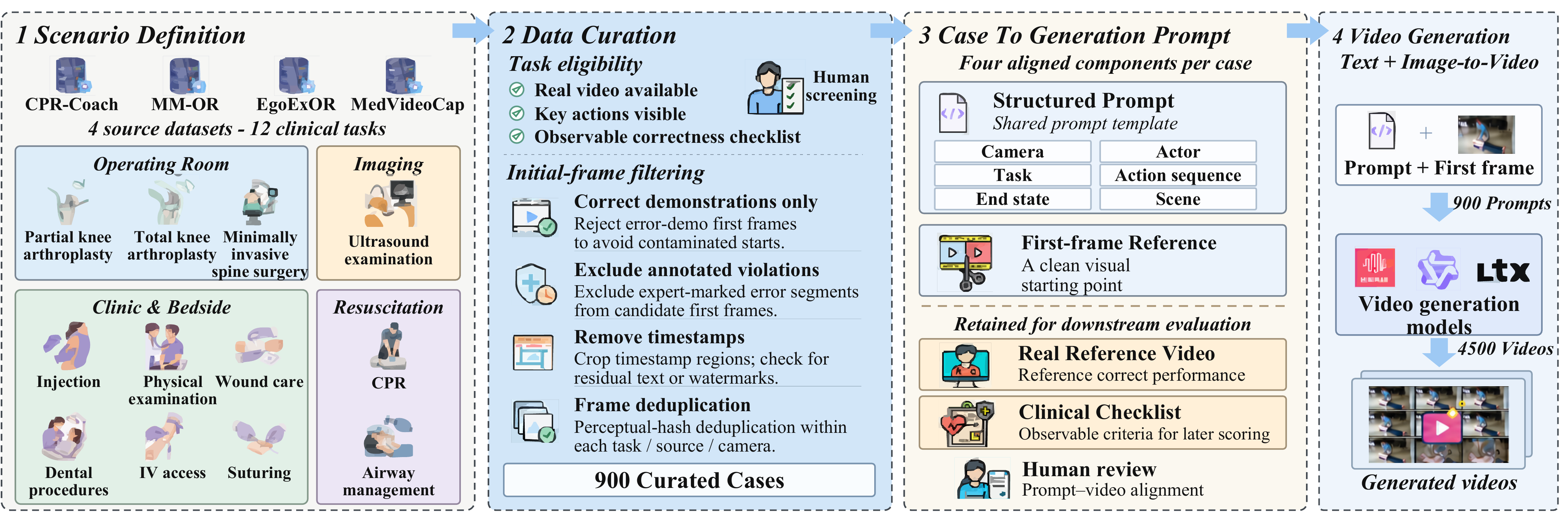}
\end{center}
\caption{\textbf{Data construction pipeline of REMEDY.}
(1)~Definition of four medical scenarios and selection of source datasets.
(2)~Screening, frame cleaning, and de-duplication yield 900 cases.
(3)~Prompt construction and human review align each case with a first frame,
reference video, and clinical checklist.
(4)~Five contemporary open-source TI2V models generate 4{,}500 videos from
the prompts and first frames.}
\label{fig:pipeline}
\end{figure}

To ensure benchmark diversity and reliable results, we use a four-stage
construction pipeline, as shown in Figure~\ref{fig:pipeline}.

\noindent\textbf{Scenario definition.}
We first define four major medical scenarios: operating room, imaging,
clinic and bedside, and resuscitation, subdivided into twelve clinical tasks.
This taxonomy defines the benchmark's scope and guides source selection,
case screening, and task-level evaluation. To cover these scenarios and
tasks, we select CPR-Coach, MM-OR, EgoExOR, and MedVideoCap and collect the
corresponding real clinical videos.

\noindent\textbf{Data curation.}
Through manual screening, we retain cases with available real videos,
visible key actions, and observable correctness criteria. We then select
first frames, excluding those from incorrect demonstrations or segments
with expert-annotated violations. We crop timestamp regions
and check for residual text or watermarks to keep procedure-irrelevant
recording dates and times out of the inputs to generators and judges.
We then remove near-duplicate candidate first frames within each task,
source, and camera group, yielding 900 cases. Source-specific screening
rules are given in Appendix~\ref{app:provenance}.

\noindent\textbf{From cases to generation prompts.}
For each curated case, we use a structured template to describe the shot,
subject, action, motion, end state, scene, and consistency requirements.
The prompt explicitly requires correct execution to professional standards.
Each case has four components: a structured prompt, a real first frame,
a real reference video, and a clinical checklist. Human review checks
that the prompt matches the video content. The first frame and prompt are
generation inputs. The real reference video supports
task-specific checklist design and motion comparisons. The checklist
records observable requirements for subsequent scoring.

\noindent\textbf{Video generation.}
We provide the first frames and prompts of all 900 cases to five contemporary
open-source TI2V models from three model families. Each model generates one video per
case, producing 4{,}500 videos in total. All models use the same first frames
and prompts to support comparison under the same task conditions.

\subsection{Benchmark statistics}
\label{sec:statistics}

\begin{figure}[htb]
\begin{center}
\includegraphics[width=\linewidth]{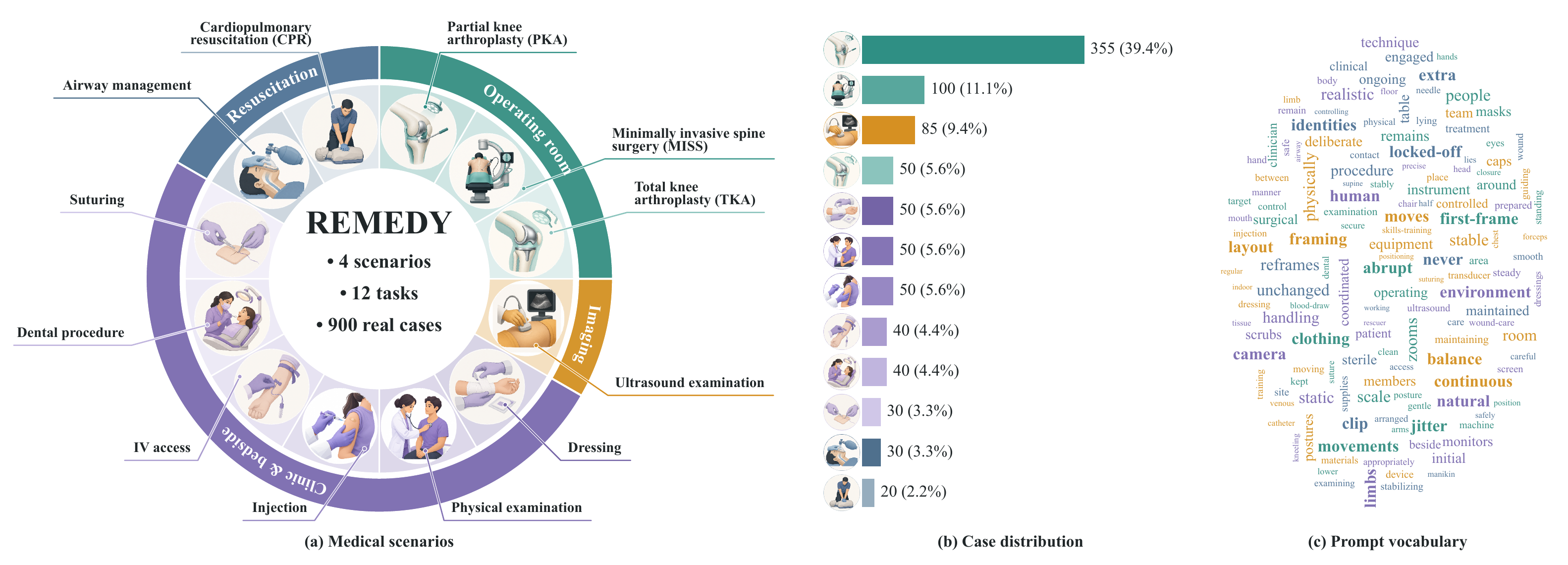}
\end{center}
\caption{\textbf{Dataset statistics of REMEDY.}
(a)~Taxonomy of twelve tasks across four medical scenarios.
(b)~Task distribution over 900 cases; bar length indicates case count, and icons correspond to (a).
(c)~Prompt word cloud, with word size indicating case frequency.}
\label{fig:overview}
\end{figure}

Figure~\ref{fig:overview} shows REMEDY's task taxonomy, case distribution,
and prompt vocabulary. REMEDY contains 900 real clinical cases.
Operating room includes 505 cases of Partial Knee Arthroplasty (PKA), Total
Knee Arthroplasty (TKA), and Minimally Invasive Spine Surgery (MISS).
Imaging contains 85 Ultrasound cases. Clinic and bedside contains 260 cases
of Injection, Physical examination, Wound care, Dental procedures,
Intra-Venous (IV) access, and Suturing. Resuscitation includes 50 cases of
Cardio-Pulmonary Resuscitation (CPR) and Airway management.
Operating-room cases account for a larger share, particularly PKA at 39.4\%.
This partly reflects the complexity of these procedures and their multiple
steps: a single surgical recording can provide clip-level cases of different
actions. Clinic and bedside covers a wider variety of
procedures and is therefore divided into six tasks to capture this diversity.

\subsection{Evaluation framework}
\label{sec:rubric}

Beyond visual fidelity and motion smoothness, REMEDY assesses the quality
of clinical execution in generated demonstrations. The framework comprises
four dimensions and twelve metrics, as summarized in Table~\ref{tab:taxonomy}.

\begin{table}[htb]
\caption{\textbf{Hierarchical evaluation framework of REMEDY.}
The framework is organized into four major dimensions: clinical action
following, clinical profiles, video quality, and motion quality, with
12 metrics for assessing generated medical teaching demonstrations.}
\label{tab:taxonomy}
\centering
\scriptsize
\setlength{\tabcolsep}{3pt}
\renewcommand{\arraystretch}{1.0}
\begin{tabular*}{\linewidth}{@{\extracolsep{\fill}}llp{0.5\linewidth}@{}}
\toprule
\bf Dimension & \bf Metric & \bf Evaluation focus \\
\midrule
\multirow{2}{*}{\shortstack[l]{Clinical action\\following}} & Strict clinical success & Whether every task-essential atom passes with no critical failure. \\
\cmidrule(lr){2-3}
 & Clinical Atom Score (CAS) & Mean value of the task's atoms on 0-100. \\
\midrule
\multirow{4}{*}{\shortstack[l]{Clinical\\profiles}} & Clinical technique (1-5) & Correctness of the visible technique for the requested procedure. \\
\cmidrule(lr){2-3}
 & Biomechanics (1-5) & Physical plausibility of body motion and clinician-patient contact. \\
\cmidrule(lr){2-3}
 & Professionalism (1-5) & Professional conduct, including step order and asepsis where required. \\
\cmidrule(lr){2-3}
 & Instruction following (1-5) & Adherence to the prompt's action and camera instructions. \\
\midrule
\multirow{4}{*}{\shortstack[l]{Video\\quality}} & Technical continuity & Black-frame, repetition, freeze, and flash penalties. \\
\cmidrule(lr){2-3}
 & Visual aesthetics & Frame-level aesthetic quality of the rendered scene. \\
\cmidrule(lr){2-3}
 & Subject consistency & Preservation of subject identity across frames. \\
\cmidrule(lr){2-3}
 & Background consistency & Preservation of scene background across frames. \\
\midrule
\multirow{2}{*}{\shortstack[l]{Motion\\quality}} & PoseAct & Similarity of visible body motion to the real reference, on 0-100. \\
\cmidrule(lr){2-3}
 & PoseCov & Fraction of clips with a valid generated/reference track pair. \\
\bottomrule
\end{tabular*}
\end{table}

\noindent\textbf{Clinical action following.}
Strict clinical success and the Clinical Atom Score (CAS) measure completion
of the required key actions in a clip. We use real reference videos to
develop task-specific checklists. During primary assessment, judges receive
these checklists but not the real reference videos. Given the prompt,
first frame, and sampled generated
frames, the judge returns a decision for each item, supporting frame
identifiers, and a brief description of the visible facts. Pass, partial,
and fail receive 1, 0.5, and 0, respectively. The score for an action or
effect cannot exceed the score of any required prerequisite. For example,
effective chest compressions require evidence of hand-chest contact and
chest response; hand motion alone is insufficient. After these constraints,
CAS is the mean of the corresponding clinical checklist scores multiplied
by 100, capturing different degrees of action completion. Strict clinical
success is the proportion of videos in which every essential item passes
and no critical failure occurs.

\noindent\textbf{Clinical profiles.}
Four metrics describe execution quality: clinical technique, biomechanics,
professionalism, and instruction following. Clinical technique assesses
whether the procedure is performed correctly, while biomechanics assesses
the plausibility of body motion and clinician-patient contact.
Professionalism checks step order and asepsis where applicable; instruction
following checks compliance with the prompt's action and camera requirements.
Each metric is computed from its corresponding checklist items. We multiply
their mean score by four and add one to obtain a score from 1 to 5. Critical
failures cap the affected metrics so that other satisfied items cannot mask
missing essential actions. Appendix~\ref{app:metrics} provides example
checklist items, evidence requirements, and scoring rules.

\noindent\textbf{Video quality.}
Following VBench and StreamAV-Bench~\citep{vbench2024,streamavbench2026},
we use four automatic metrics to assess visual appearance and stability over time:
technical continuity, visual aesthetics, subject consistency, and background
consistency. Technical continuity checks the full clip for black frames,
repeated frames, freezes, and flashes. Visual aesthetics uses an aesthetic
prediction model to score sampled frames. Subject and background consistency
compare visual features across frames to measure how well the subject and
background are preserved, respectively.

\noindent\textbf{Motion quality.}
PoseAct measures the similarity of body motion between generated and real
reference videos, while PoseCov reports the proportion of cases available
for this comparison. We extract 2D human keypoint trajectories from both
videos, smooth them, and normalize them by torso size. We then compare joint
trajectories, angles, velocities, and motion energy to compute PoseAct on a
0-100 scale. For valid generated/reference track pairs, we first average
PoseAct within each task, then equally across the twelve tasks. PoseCov is
the fraction of cases with valid track pairs within each task, averaged
equally across tasks to report the coverage of motion evaluation.

\section{Experiments}
\label{sec:experiments}
\subsection{Experimental setup}
\label{sec:exp_setup}

\noindent\textbf{Evaluated models.}
To compare the generation of subsequent actions from the same clinical
starting state, we focus on TI2V models that accept a first-frame image as
conditioning input. Specifically, we evaluate five contemporary open-source
TI2V models: Wan2.2-TI2V-5B, LTX-Video-2B, LTX-2.3, LTX-2.5, and
MiniMax-H3-FL2VA, drawn from three model families~\citep{wan2025,ltx2paper,minimaxh32026}.
These models range in size from 2B to 33B parameters.

\noindent\textbf{Implementation details.}
We run all models using their official checkpoints and inference pipelines
on NVIDIA H200 GPUs. To keep the evaluation protocol consistent, all models
receive the same 900 real first frames and task-specific prompts.
We use GPT-4o to judge whether generated videos satisfy each clinical
checklist item. On 450 human-reviewed generated videos, we compare agreement between human
raters and between GPT-4o and humans, Qwen3-VL-32B, and Gemini
(Appendix~\ref{app:cred}).

\subsection{Model performance}
\label{sec:main_results}

Table~\ref{tab:main_results} shows that none of the five evaluated
open-source models reliably completes the required key actions across all
tasks. MiniMax-H3 shows the strongest overall clinical performance among
the evaluated systems. Even so, its overall strict clinical success rate is only 28.25\% across all twelve tasks. In video quality, the LTX family has a clear
advantage in technical continuity: all three LTX models score above 90 on
this metric.

\begin{table*}[htb]
\caption{\textbf{Evaluation results on REMEDY.}
Five contemporary open-source models are compared on task-level strict clinical success and CAS (top and
middle), and clinical profiles, video quality, motion quality, and generation
speed (bottom).
Bold and underlined values indicate the best and second-best results.}
\label{tab:main_results}
\scriptsize
\renewcommand{\arraystretch}{0.90}
\setlength{\tabcolsep}{3pt}

\begin{tabular*}{\textwidth}{@{\extracolsep{\fill}}l*{12}{c}@{}}
\toprule
 & \multicolumn{12}{c}{\bf Strict clinical success (\%) $\uparrow$} \\
\cmidrule(lr){2-13}
\bf Model & \bf PKA & \bf TKA & \bf MISS & \bf US & \bf Inj. & \bf Exam &
\bf Wound & \bf Dent. & \bf IV & \bf Sut. & \bf CPR & \bf Air \\
\midrule
MiniMax-H3-FL2VA & \bf 25.07 & \bf 6.00 & \bf 13.00 & 9.52 & \bf 16.00 & \bf 74.00 & \bf 68.00 & \bf 62.50 & \bf 50.00 & \bf 6.67 & \bf 15.00 & \bf 40.00 \\
Wan2.2-TI2V-5B   & \underline{21.97} & \underline{2.00} & 5.05 & \underline{13.10} & \underline{12.00} & 44.00 & \underline{34.00} & \underline{17.50} & \underline{7.50} & 0.00 & \underline{5.00} & 16.67 \\
LTX-2.5     & 12.11 & \bf 6.00 & \underline{10.00} & 4.71 & 6.00 & \underline{50.00} & 16.00 & 15.00 & 2.50 & 0.00 & 0.00 & \underline{33.33} \\
LTX-2.3v     & 0.85 & 0.00 & 0.00 & \bf 14.29 & 4.00 & 36.00 & 6.00 & 2.50 & 2.50 & \underline{3.45} & 0.00 & 6.67 \\
LTX-Video-2B     & 1.41 & 0.00 & 0.00 & 3.53 & 4.00 & 34.00 & 6.00 & 10.00 & 0.00 & 0.00 & 0.00 & 3.33 \\
\bottomrule
\end{tabular*}

\vspace{4pt}
\begin{tabular*}{\textwidth}{@{\extracolsep{\fill}}l*{12}{c}@{}}
\toprule
&\multicolumn{12}{c}{\bf CAS $\uparrow$} \\
\cmidrule(lr){2-13}
\bf Model & \bf PKA & \bf TKA & \bf MISS & \bf US & \bf Inj. & \bf Exam &
\bf Wound & \bf Dent. & \bf IV & \bf Sut. & \bf CPR & \bf Air \\
\midrule
MiniMax-H3-FL2VA & \bf 52.31 & \bf 39.80 & \bf 48.10 & \bf 35.60 & \bf 71.67 & \bf 85.00 & \bf 83.50 & \bf 80.25 & \bf 63.12 & \bf 28.61 & \bf 51.25 & \bf 71.33 \\
Wan2.2-TI2V-5B   & \underline{46.39} & 21.40 & \underline{41.52} & 21.31 & \underline{57.50} & \underline{73.25} & \underline{66.17} & 43.00 & 20.00 & 15.83 & 29.38 & 53.67 \\
LTX-2.5     & 44.42 & \underline{36.40} & 34.00 & \underline{34.00} & 53.17 & 71.25 & 46.00 & \underline{46.25} & \underline{26.67} & \underline{20.28} & 20.00 & \underline{65.67} \\
LTX-2.3     & 10.48 & 8.20 & 10.10 & 27.74 & 18.50 & 46.00 & 15.00 & 38.00 & 15.62 & 16.95 & \underline{33.44} & 43.00 \\
LTX-Video-2B     & 15.52 & 9.80 & 16.10 & 10.71 & 39.33 & 50.50 & 30.17 & 26.25 & 10.62 & 7.78 & 19.06 & 29.33 \\
\bottomrule
\end{tabular*}

\vspace{4pt}
\begin{tabular*}{\textwidth}{@{\extracolsep{\fill}}l*{11}{c}@{}}
\toprule
\bf Model & \multicolumn{4}{c}{\bf Clinical profiles} &
\multicolumn{4}{c}{\bf Video quality} &
\multicolumn{2}{c}{\bf Motion quality} & \bf Efficiency \\
\cmidrule(lr){2-5}\cmidrule(lr){6-9}\cmidrule(lr){10-11}\cmidrule(lr){12-12}
& \bf Clin. $\uparrow$ & \bf Biomech. $\uparrow$ & \bf Prof. $\uparrow$ &
\bf Instr. $\uparrow$ & \bf Tech. $\uparrow$ & \bf Aesth. $\uparrow$ &
\bf SC $\uparrow$ & \bf BC $\uparrow$ & \bf PoseAct $\uparrow$ &
\bf PoseCov $\uparrow$ & \bf FPS $\uparrow$ \\
\midrule
MiniMax-H3-FL2VA & \bf 2.78 & \bf 3.17 & \bf 3.16 & \bf 2.86 & 70.48 & \bf 0.488 & \underline{0.932} & \underline{0.960} & 43.99 & \underline{0.644} & 0.288 \\
Wan2.2-TI2V-5B   & \underline{1.88} & 2.51 & \underline{2.41} & \underline{1.94} & 88.19 & \underline{0.476} & \bf 0.940 & \bf 0.961 & \underline{44.65} & 0.636 & \underline{0.754} \\
LTX-2.5     & 1.76 & \underline{2.55} & 2.21 & 1.64 & \bf 98.70 & 0.461 & 0.806 & 0.901 & 37.58 & \bf 0.657 & 0.645 \\
LTX-2.3     & 1.42 & 1.82 & 1.78 & 1.35 & \underline{95.68} & 0.474 & 0.782 & 0.890 & 32.62 & 0.640 & 0.649 \\
LTX-Video-2B     & 1.37 & 1.74 & 1.81 & 1.41 & 92.17 & 0.443 & 0.886 & 0.930 & \bf 44.77 & 0.621 & \bf 8.975 \\
\bottomrule
\end{tabular*}
\end{table*}

The clinical results reveal a gap between satisfying some requirements and
completing all essential actions. Higher CAS does not always coincide with
higher strict success. Specifically, the former measures average checklist
satisfaction, whereas the latter requires every essential item to pass without a critical
failure. Improvements in some actions may therefore be insufficient for
more videos to pass the complete checklist. The clinical profiles also
reveal weaknesses in execution. Even the highest clinical technique and
instruction-following scores are only 2.78 and 2.86 out of 5, respectively,
indicating persistent difficulty in performing procedures correctly and
following the requested actions. Strengths in video quality do not consistently align with clinical
performance. Technical continuity reaches 98.70, and background consistency
ranges from 0.890 to 0.961 across models, yet the leaders on these metrics
are not the strongest clinically. This contrast separates the ability to
reduce technical artifacts and preserve people and backgrounds from the
ability to perform the requested procedure. Stable visual content cannot
replace checks of whether actions are correct and task requirements are met.


Motion similarity likewise does not directly indicate clinical correctness.
The model with the highest PoseAct ranks
last on both aggregate clinical measures, showing that similar coarse body
motion need not imply correct fine hand manipulation or tissue response.
Valid track coverage is only 62.1\%--65.7\% across models, so motion
comparisons are also limited to clips with usable tracks. Under the
model-specific output settings, higher generation throughput does not
necessarily correspond to better clinical performance either. Motion and
efficiency metrics therefore provide complementary information rather than
substitutes for procedural correctness. Model profiles appear in
Appendix~\ref{app:leaderboard}.

\subsection{Analysis and Findings}
\label{sec:analysis_findings}

\begin{figure}[htb]
\centering
\includegraphics[width=\linewidth]{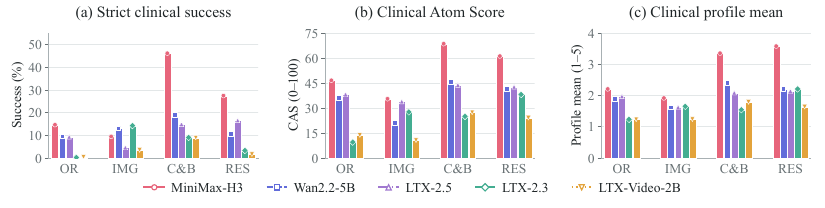}
\caption{\textbf{Scenario-level clinical performance.}
(a)~Strict clinical success (\%); (b)~CAS (0-100); (c)~mean clinical profile
score (1-5).}
\label{fig:scenario_balanced}
\end{figure}

Figure~\ref{fig:scenario_balanced} compares clinical performance across the
four medical scenarios, weighting tasks equally within each scenario.
Strict clinical success rates in operating-room and imaging scenarios remain
below 15\%, substantially lower overall than in clinic and bedside and
resuscitation scenarios. Even H3, the
overall clinical leader, achieves only 14.69\% and 9.52\% in these two
scenarios, respectively.

\begin{figure}[htb]
\begin{center}
\includegraphics[width=\linewidth]{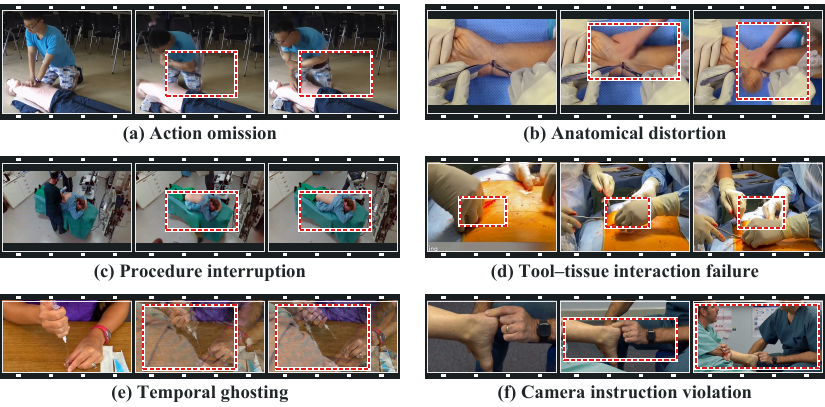}
\end{center}
\caption{\textbf{Typical failure modes in generated medical videos.}
(a)~Action omission; (b)~anatomical distortion; (c)~procedure interruption;
(d)~tool-tissue interaction failure; (e)~temporal ghosting;
(f)~camera instruction violation. Each example shows three temporally
ordered frames. Red dashed boxes highlight the failures.}
\label{fig:qualitative}
\end{figure}

As illustrated in Figure~\ref{fig:qualitative}, we identify
six common failure modes in generated medical demonstrations.
Action omission (a) occurs when a required key action is absent or replaced
by unrelated movement. Anatomical distortion (b) produces deformed body
structures or extra limbs. Procedure interruption (c) occurs when an action
begins but fails to continue or progress as required. Tool-tissue interaction
failure (d) occurs when tools move without visible evidence of the required
contact or corresponding tissue change. Temporal ghosting (e) produces
overlapping residual images of moving bodies or objects, making actions and
contact relations difficult to discern. Camera instruction violation (f)
occurs when camera motion or reframing departs from the prompt, changing
which parts of the procedure remain visible. Overall, these observations
highlight bottlenecks in sustained execution of required actions, anatomical
stability, tool-tissue interaction responses, temporal consistency, and
camera instruction following. Future improvements should therefore go
beyond visual quality to sustain key actions, produce the visible
interaction effects required by the task, and maintain a clear view that
follows the camera instructions.

\subsection{Human Alignment}
\label{sec:human_alignment}

To provide a human reference for clinical action assessment, six raters
ranked the clinical credibility of 600 generated videos, with 100 videos assigned to
each rater. Videos were grouped by shared input conditions, with one output
from each of the five models per group. Using the corresponding real
reference video, raters ranked the five anonymized outputs within each
group, with ties allowed.
To reduce scoring differences arising from how raters interpret metric
definitions, we use relative ranking to focus the assessment on overall
clinical credibility within the same task. This simplifies the review
process and reduces ambiguity from differing standards for absolute
scores, keeping attention on which outputs more closely resemble credible
clinical demonstrations.
Table~\ref{tab:human_alignment} reports the mean ranking scores, with
MiniMax-H3 receiving the highest mean.

\begin{table}[htb]
\caption{\textbf{Human rankings of clinical credibility.}
Six raters assess 100 generated videos each, totaling 600 videos, and rank
the five model outputs within each group sharing the same input conditions.
Scores are $6-\mathrm{rank}$ (1-5, higher is better). Bold indicates the highest mean.}
\label{tab:human_alignment}
\centering
\small
\setlength{\tabcolsep}{4.0pt}
\begin{tabular*}{\linewidth}{@{\extracolsep{\fill}}l*{5}{c}@{}}
\toprule
& \bf MiniMax-H3 & \bf Wan2.2-5B & \bf LTX-2.5 & \bf LTX-2.3 & \bf LTX-2B \\
\midrule
Ranking score (1-5) $\uparrow$ & \bf 4.58 & 3.29 & 2.50 & 2.63 & 1.75 \\
\bottomrule
\end{tabular*}
\end{table}

\section{Related Work}
\label{sec:related}


\subsection{Medical Video Synthesis}
\label{sec:rw-medvid}

Early medical video generators, including endoscopy simulators and
general-purpose biomedical models, are evaluated primarily with
distributional metrics such as FVD~\citep{endora2024,bora2024,fvd2018}.
MedGen fine-tunes a text-to-video model on 55K captioned medical clips
and shows that general-purpose generators produce anatomical and
instrument errors~\citep{medgen2025}. SurgVeo benchmarks zero-shot
video continuation for endoscopic surgery~\citep{surgveo2025}. Both
lines of work focus on in-body endoscopic views. REMEDY targets a
complementary setting, medical teaching demonstrations in which the
performer is visible, that additionally requires the generated video
to depict correct body mechanics and procedural actions simultaneously.

\subsection{Benchmarks for Video Generation}
\label{sec:rw-bench}

Evaluation of generated video has developed along two lines. The first
decomposes quality into per-dimension metrics such as appearance,
temporal coherence, and prompt alignment, and extends them to
compositional prompts, faithfulness beyond surface appearance, and
learned
scores~\citep{vbench2024,evalcrafter2024,t2vcompbench2024,vbench2_2025,videoscore2024}.
MLLM-based benchmarks complement this approach by having a judge model
answer rubric questions per
clip~\citep{videobench2025,ui2vbench2025}. Educational evaluations
prioritize conceptual correctness over perceptual
realism~\citep{eduvqa2026}, biomechanical benchmarks verify
reconstructed 3D human motion against anatomical constraints under
single-person, unoccluded
assumptions~\citep{humanscore2026}, and large-scale human studies
collect action-quality ratings~\citep{gaia2024}. The second line tests
physical plausibility, i.e.\ whether generated events obey physical
rules~\citep{videophy2024,phygenbench2024,physicsiq2025}. In the
medical domain, Med-VBench applies general-purpose metrics to medical
prompts and adds a small-scale clinician preference
study~\citep{medgen2025}. However, none of the above benchmarks
specifies which visible actions a particular clinical procedure
requires, and therefore none can determine whether a generated clip
performs all required key actions for that task. REMEDY fills this gap
by defining task-specific checklists and extending physical-plausibility
evaluation to clinical biomechanics and procedural motion.

\subsection{Model-Based Judge Reliability}
\label{sec:rw-judge}

A growing line of research documents reliability issues in model-based
evaluation~\citep{zheng2023judging,mllmjudge2024}. Video LLMs infer
human motion from scene-level co-occurrence rather than visible
evidence~\citep{mohallbench2026}. VLM judges produce reliable rankings
yet assign unreliable absolute scores~\citep{kumar2026rank}, and
exhibit fixed tendencies insensitive to
context~\citep{alloula2026rigid}. Text-only LLM judges can express
preferences even between empty, whitespace-only, or identical
answers~\citep{usami2026darkcurrent}. In medical text, LLM judges and
clinicians systematically diverge in reasoning even when they agree on
the final verdict~\citep{delucia2026verdict}. Shortcut exploitation
can inflate benchmark scores without the intended
capability~\citep{ren2026shortcut}, and no tested family of AI-video
detectors generalizes across generation
sources~\citep{rabench2026}. These findings motivate REMEDY's
evaluation protocol: we measure inter-human, human-judge, and
cross-judge clinical-score agreement on 450 human-reviewed generated
videos, and complement the automatic results with a separate
human-ranking study.

\section{Conclusion}
\label{sec:conclusion}

We introduced REMEDY, a benchmark that defines the key actions
required by each clinical procedure and compares automatic clinical
judgments with human judgments. Evaluating five open-source TI2V
models reveals both promise and limitations: the strongest model,
MiniMax-H3, leads most tasks in automatic clinical scores and
aggregate human rankings (Table~\ref{tab:human_alignment}), showing
that general-purpose video generation progress can translate into
medically meaningful actions. However, strict success remains low
across most procedures and near zero for some, indicating a
considerable gap between recognizable clinical scenes and clips
suitable for direct use as teaching material. Our conclusions are
scoped to the five evaluated models with GPT-4o as the automatic
judge; score-agreement validation on 450 human-reviewed clips shows
that binary strict-success judgments largely agree while fine-grained
clinical-profile scores still diverge. More broadly, by making required actions explicit and reusable,
REMEDY can track generator progress over time and supply measurable
targets for future generators and world models aimed at medical education.

\subsection*{AI use statement}
We used AI tools to assist with writing and polishing, literature retrieval,
and generating prompt text. Specifically, AI assisted with Chinese--English
translation, language polishing, related-work discovery, and reference
checking. In dataset construction, generative AI assisted in drafting video
generation prompts, and human review checked their textual quality and
consistency with the source videos. The authors take full responsibility
for developing and carrying out the research plan, as well as the final
research decisions, interpretations, references, and manuscript content.

\subsection*{Ethics statement}
REMEDY evaluates generated educational media and does not provide clinical
guidance or measure patient outcomes. Source datasets are used under their
respective research terms. The CPR source depicts manikin-based training;
operating-room sources are established research datasets whose original
custodians govern consent and access. The benchmark release does not
redistribute restricted source video: it provides derived manifests,
annotations, and source pointers where required by the upstream terms.
Generated clips are marked as synthetic and intended for research evaluation
only. They should not be used for instruction without qualified clinical
review because visually plausible outputs may contain unsafe technique.

\subsection*{Reproducibility statement}
Upon acceptance, we will publicly release the REMEDY benchmark data and
the code for reproducing the experiments and constructing the benchmark.
The release will include prompts and source-video identifiers for all 900
cases, 4{,}500 generated videos, clinical checklists and scoring rubrics,
per-video evaluation records and generation configurations, and scripts for
data construction, video generation, evaluation, and analysis. The original
clinical videos must be obtained from the dataset providers under their
respective licenses and will not be redistributed.

\bibliography{iclr2027_conference}
\bibliographystyle{iclr2027_conference}

\appendix

\raggedbottom
\section{Data curation and prompt design}
\label{app:provenance}
\label{app:systems}
The 900 cases come from four datasets. CPR-Coach provides 20 cases; we
retain only correct demonstrations and remove recorder timestamps.
MM-OR provides 405 cases, comprising 355 partial and 50 total knee
arthroplasty cases; we exclude frames covered by sterility-breach annotations.
EgoExOR provides 185 cases: 100 minimally invasive spine surgery and 85
ultrasound cases. MedVideoCap provides 290 cases covering injection,
physical examination, dressing, dental procedures, venous access, airway
management, and suturing. Across all sources, we remove near-duplicate
frames within each (task, source, camera) group.

The generation template below is filled with task-specific subject, action,
end-state, and scene descriptions. Scene and consistency requirements share
the final paragraph.

\noindent\fbox{\begin{minipage}{\dimexpr\linewidth-2\fboxsep-2\fboxrule\relax}
\small
\textbf{Generation prompt template}\par\smallskip
\setlength{\tabcolsep}{3pt}
\begin{tabular}{@{}>{\raggedright\arraybackslash}p{.13\linewidth}>{\raggedright\arraybackslash}p{.83\linewidth}@{}}
Shot & A \{duration\}-second locked-off static shot; the camera never moves, zooms, or reframes. \\[4pt]
Subject & \{subject\}, exactly as positioned in the first frame. \\[4pt]
Action & Starting from the first-frame positions, \{action\}. \\[4pt]
Motion & All human motion is natural, continuous, physically realistic: stable balance, no jitter, no extra limbs, no abrupt movements. \\[4pt]
End State & At the end of the clip, \{end\}. \\[4pt]
Scene & \{scene\}: keep the environment, people's identities, clothing, layout, framing and scale unchanged throughout. \\[4pt]
\end{tabular}
\end{minipage}}

\medskip
\noindent\fbox{\begin{minipage}{\dimexpr\linewidth-2\fboxsep-2\fboxrule\relax}
\small
\textbf{Example task-specific action: CPR}\par\smallskip
Starting from the first-frame positions, the rescuer performs continuous
chest compressions on the patient, correctly to professional standard:
hands on the lower half of the breastbone, arms straight and vertical,
a steady regular compression rhythm.
\end{minipage}}

\section{Clinical checklists and scoring}
\label{app:metrics}
\label{app:rubric}
Each video is judged with three shared questions and its task-specific
checklist. The questions below are reproduced from the evaluation rubric. Reference
videos inform checklist design but are not shown during primary scoring.


\begin{table}[htb]
\caption{\textbf{Shared checks for generated videos.}}
\label{tab:shared_checks_supp}
\centering\small
\setlength{\tabcolsep}{4pt}
\renewcommand{\arraystretch}{1.4}
\begin{tabular}{@{}>{\raggedright\arraybackslash}p{.54\linewidth}>{\raggedright\arraybackslash}p{.44\linewidth}@{}}
\toprule
\textbf{Question} & \textbf{Required evidence} \\
\midrule
Is the requested procedure visibly performed with task-relevant motion, rather than only a pose, hand jitter, or unrelated movement? & Cite separated labelled frames showing the task begins and continues or reaches a later state. Hidden or omitted action is FAIL. \\
\hline
Does the camera remain locked off, without viewpoint motion, zoom, reframing, or crop drift beyond negligible encoding jitter? & Compare stable background landmarks near the beginning, middle, and end. \\
\hline
Are the visible people, anatomy, clothing, equipment, layout, framing, and scale preserved from the conditioning frame? & Compare the conditioning frame with labelled generated frames and name material drift. \\
\bottomrule
\end{tabular}
\end{table}

\begin{table}[htb]
\caption{\textbf{Example clinical questions for all twelve tasks.}
One question per task is shown; scoring uses the full checklist.}
\label{tab:task_checks_supp}
\centering\small
\setlength{\tabcolsep}{4pt}
\renewcommand{\arraystretch}{1.4}
\begin{tabular}{@{}>{\raggedright\arraybackslash}p{.25\linewidth}>{\raggedright\arraybackslash}p{.73\linewidth}@{}}
\toprule
\textbf{Task} & \textbf{Task-specific question} \\
\midrule
Partial knee arthroplasty & Are visible instruments handled deliberately without implausible detachment or penetration? \\
\hline
Total knee arthroplasty & Does visible team activity sustain or advance the ongoing procedure? \\
\hline
Minimally invasive spine surgery & Is visible imaging or spine equipment positioned and handled in a controlled way? \\
\hline
Ultrasound & Does the transducer maintain continuous plausible contact with the intended body surface? \\
\hline
Injection & Does the visible sequence progress through a coherent injection action? \\
\hline
Physical examination & Are deliberate examination maneuvers visibly performed rather than a static touch? \\
\hline
Dressing & Does the wound or dressing reach a visibly advanced controlled state? \\
\hline
Dental procedure & Is instrument contact deliberate without detachment, penetration, or anatomy distortion? \\
\hline
Venous access & Does the device remain coupled to the site without sliding, hovering, or penetration artifacts? \\
\hline
Suturing & Does the needle visibly engage or pass through the tissue rather than hover? \\
\hline
CPR & Does the chest target itself visibly depress under the applied force? \\
\hline
Airway management & Does the device maintain controlled plausible contact or seal? \\
\bottomrule
\end{tabular}
\end{table}

\begin{table}[htb]
\caption{\textbf{Implementation of video and motion metrics.}}
\label{tab:metric_implementation_supp}
\centering\small
\setlength{\tabcolsep}{4pt}
\renewcommand{\arraystretch}{1.4}
\begin{tabular}{@{}>{\raggedright\arraybackslash}p{.26\linewidth}>{\raggedright\arraybackslash}p{.72\linewidth}@{}}
\toprule
\textbf{Metric} & \textbf{Features and aggregation} \\
\midrule
Visual aesthetics & LAION aesthetic head~\citep{laion5b2022,vbench2024} over
CLIP ViT-L/14 features~\citep{clip2021};
32 uniformly sampled frames. \\
\hline
Subject consistency & DINO ViT-B/16 features~\citep{dino2021};
32 uniformly sampled frames. \\
\hline
Background consistency & CLIP ViT-B/32 features; 32 uniformly sampled frames. \\
\hline
Technical continuity & Full-clip penalties for black frames, repetition,
freezes, and flashes. \\
\hline
PoseAct & YOLO11m-Pose body tracks~\citep{yolo11ultralytics};
smoothed, torso-normalized trajectories,
joint angles, velocities, and motion energy compared with the real reference.
Average valid pairs within tasks, then equally across twelve tasks. \\
\hline
PoseCov & Task-balanced fraction of valid generated/reference track pairs;
failed tracks are not imputed. \\
\hline
Generation FPS & Generated frame count divided by measured generation time,
not playback frame rate. \\
\bottomrule
\end{tabular}
\end{table}
\label{app:motionphys}

\section{Judge validation}
\label{app:cred}
\subsection{Validation scope}
\label{app:probes}
We assess agreement on strict clinical success and the four clinical-profile
metrics using 450 human-reviewed generated videos. Separately, one real reference video per
task checks whether real demonstrations receive reasonably high scores.
Both analyses are distinct from the human-ranking study of 600 generated
videos in the main text.
The main evaluation uses fixed checklists and GPT-4o judgments for
4,500 generated clips.
\label{app:motion_control}

\subsection{Score agreement on generated videos}
\label{app:score_agreement}
We assessed score agreement on 450 human-reviewed generated videos.
Table~\ref{tab:validation_protocol_supp} compares human raters with one
another and GPT-4o with humans, Qwen-32B, and Gemini. For strict clinical
success, we report the number of videos with different binary judgments,
rather than a success rate or mean absolute difference. For clinical
technique, biomechanics, professionalism, and instruction following, we
report Mean L1: the average absolute score difference across videos on
the original 1--5 scales.

\begin{table}[htb]
\centering\small
\caption{\textbf{Agreement of clinical scores on 450 human-reviewed generated videos.}
Strict Clinical Success reports the number of videos with disagreeing
binary judgments out of 450. The four clinical-profile columns report
Mean L1 on the original 1--5 score scales. Lower is better in all columns.}
\label{tab:validation_protocol_supp}
\setlength{\tabcolsep}{3pt}
\begin{tabular*}{\linewidth}{@{\extracolsep{\fill}}lrrrrr@{}}
\toprule
Comparison & \shortstack{Strict Clinical\\Success\\Disagreements} &
\shortstack{Clinical\\Technique} & Biomechanics & Professionalism &
\shortstack{Instruction\\Following} \\
\midrule
Inter-Human & 0 & 0.907 & 0.921 & 1.121 & 1.501 \\
GPT-4o vs. Human & 2 & 1.217 & 1.705 & 1.600 & 1.035 \\
GPT-4o vs. Qwen-32B & 4 & 1.317 & 1.045 & 1.340 & 1.515 \\
GPT-4o vs. Gemini & 1 & 1.402 & 1.410 & 1.595 & 1.627 \\
\bottomrule
\end{tabular*}
\end{table}

Across these 450 generated videos, human raters agreed on strict clinical success,
while GPT-4o differed from humans on two videos. GPT-4o differed from
Qwen-32B on four videos and from Gemini on one. Although the binary
judgments largely agree, the four clinical-profile scores still differ,
including between human raters.

\subsection{Scores on real reference videos}
\label{app:reference_agreement}
Separately, we scored real references with GPT-4o and
Qwen3-VL-32B, without preset full scores.
Both received the same clinical questions and sampled frames. Table~\ref{tab:real_reference_scores} summarizes
strict clinical success rates and mean clinical scores.
Across 62 clinical items, agreement was
87.10\%, quadratic-weighted $\kappa$ was 0.601, and CAS Spearman
correlation was 0.769. These references were used in checklist development
and mainly assess calibration near the upper end of the score range.

\begin{table}[htb]
\centering\small
\caption{\textbf{Clinical scores on real reference videos.}
Strict clinical success is reported as a percentage. CAS and clinical-profile
scores are means across the videos, on 0--100 and 1--5 scales, respectively.}
\label{tab:real_reference_scores}
\setlength{\tabcolsep}{3pt}
\begin{tabular*}{\linewidth}{@{\extracolsep{\fill}}lrrrrr@{}}
\toprule
Judge & \shortstack{Strict Clinical\\Success (\%)} & CAS &
\shortstack{Clinical\\Technique} & Biomechanics & Professionalism \\
\midrule
GPT-4o & 93.33 & 94.79 & 4.79 & 4.63 & 4.86 \\
Qwen3-VL-32B & 85.00 & 88.75 & 4.38 & 4.33 & 4.40 \\
\bottomrule
\end{tabular*}
\end{table}

\section{Supplementary model comparisons}
\label{app:leaderboard}
\begin{figure}[htb]
\centering
\includegraphics[width=\linewidth]{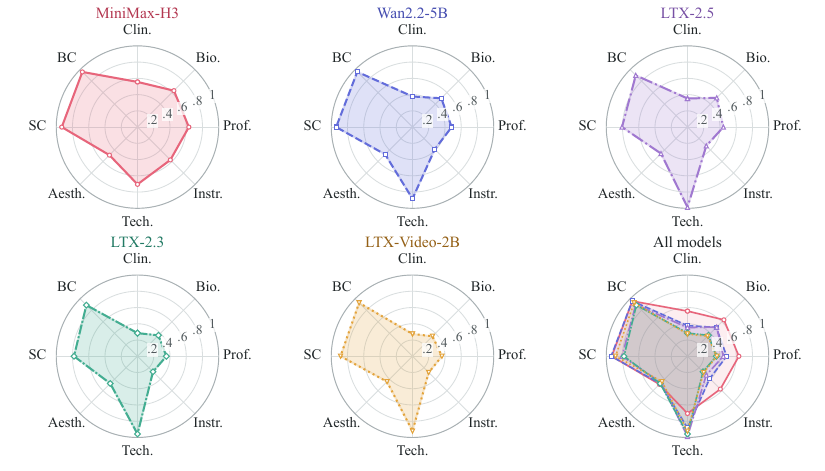}
\caption{\textbf{Model profiles across clinical and video-quality dimensions.}
Five individual radar plots and an overlay compare task-balanced scores.
Scores are averaged equally across twelve tasks and divided by their
metric's maximum to share a 0--1 scale. Clinical profiles summarize valid
judgments from the 4{,}500 generated clips.}
\label{fig:capability_profile}
\end{figure}

Figure~\ref{fig:capability_profile} summarizes the main-table results by
evaluation dimension. Scenario-level clinical results appear in
Section~\ref{sec:analysis_findings}.

\section{Additional failure examples}
\label{app:examples}
We show two additional cases for each of the six failure modes.
Each row pairs an input frame with five generated frames in temporal order.
Red marks locate the visible issue; the sentence below describes it.
Frames use fixed 4:3 crops and are resized without stretching.

\begin{figure}[htb]
\centering
\includegraphics[width=\linewidth]{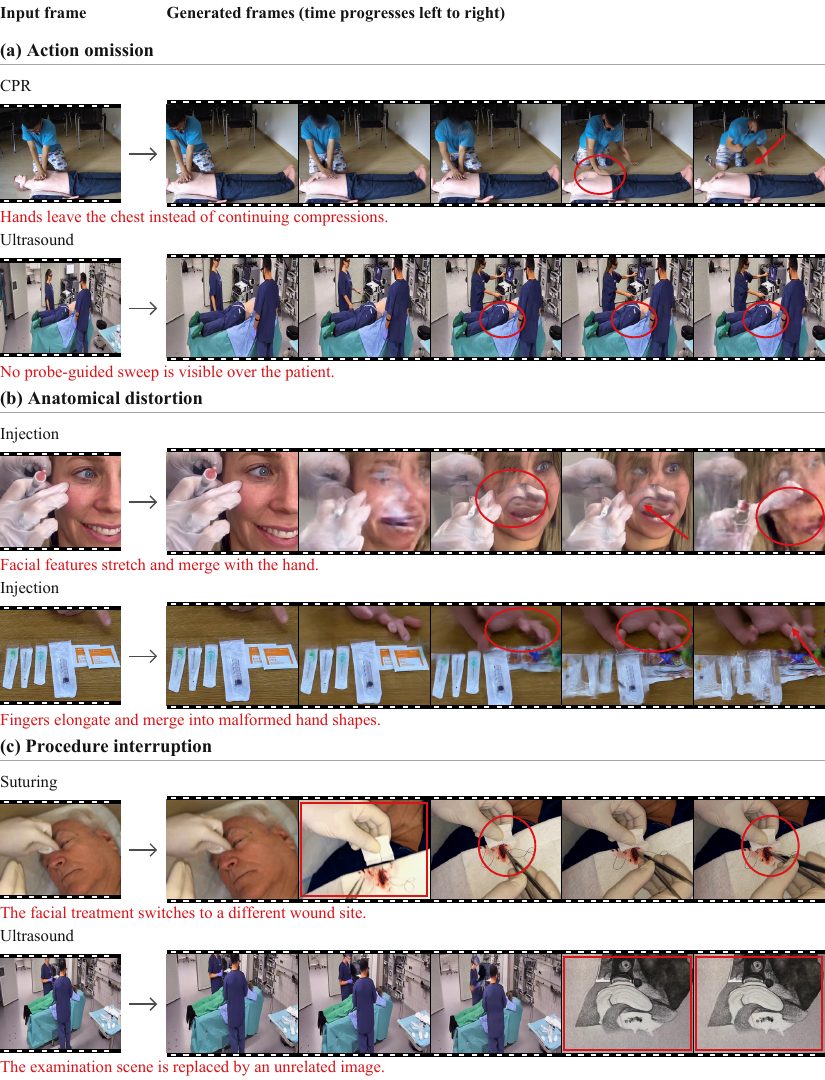}
\caption{\textbf{Action omission, anatomical distortion, and procedure interruption.}
Two examples illustrate each failure mode. Red circles, boxes, and arrows
highlight the affected regions or the change of scene.}
\label{fig:failure_examples_1}
\end{figure}

\begin{figure}[htb]
\centering
\includegraphics[width=\linewidth]{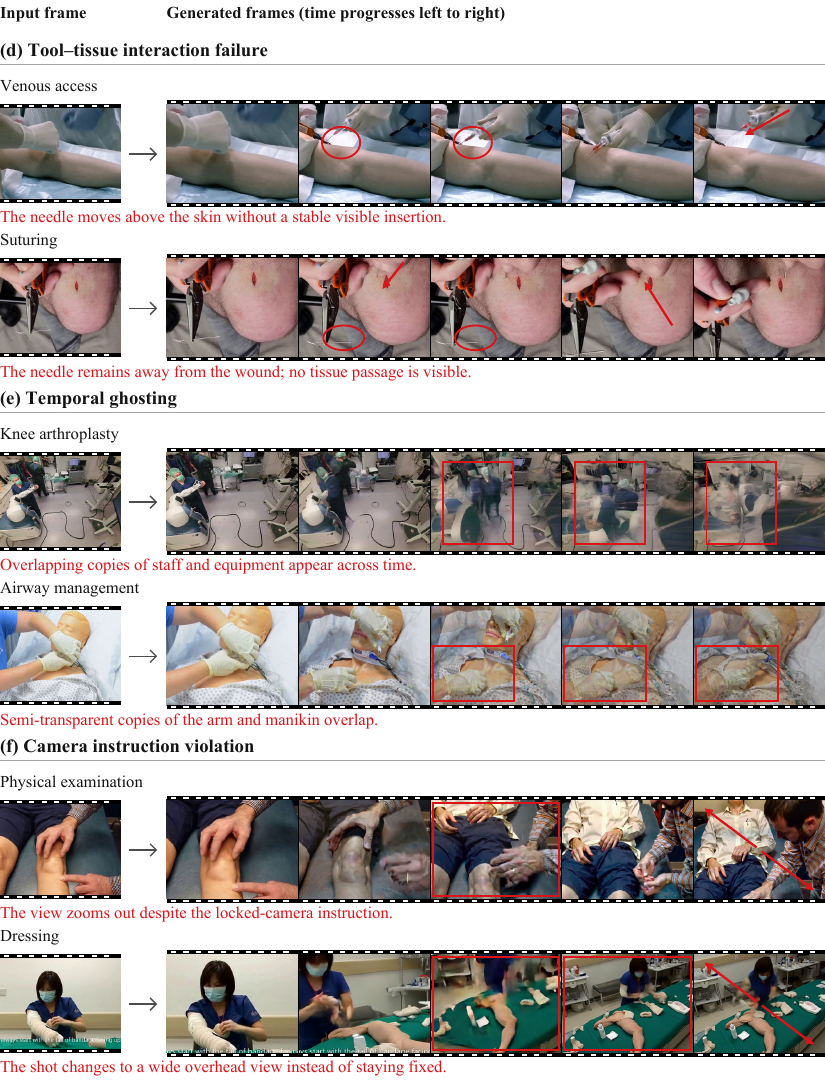}
\caption{\textbf{Tool--tissue interaction failures, temporal ghosting, and camera instruction violations.}
Two examples illustrate each failure mode. Red marks highlight the
tool--target region, overlapping silhouettes, or changes in framing
despite the fixed-camera prompt.}
\label{fig:failure_examples_2}
\end{figure}

\end{document}